\pdfoutput=1
\documentclass[11pt]{article}

\usepackage[]{ACL2023}

\usepackage{times}
\usepackage{latexsym}
\usepackage{multirow}
\usepackage{graphicx}
\usepackage{amsmath,amsfonts,amssymb}
\usepackage{booktabs, graphics}
\usepackage[normalem]{ulem}
\usepackage{pifont}% http://ctan.org/pkg/pifont
\newcommand\cmark {\textcolor{green}{\ding{52}}}
\newcommand\xmark {\textcolor{red}{\ding{55}}}

\newcommand{\cmtt}[1]{{\fontfamily{cmtt}\selectfont {#1}}}

\usepackage[T1]{fontenc}
\usepackage[utf8]{inputenc}
\usepackage{array}
\usepackage{microtype}

\usepackage{pgfplots}
\pgfplotsset{compat=1.10} 
\usepackage{filecontents}

\title{Real-World Compositional Generalization\\ with Disentangled Sequence-to-Sequence Learning }

\author{Hao Zheng \textnormal{and} Mirella Lapata\\
Institute for Language, Cognition and Computation\\
School of Informatics, University of Edinburgh\\
 10 Crichton Street, Edinburgh EH8 9AB\\
\texttt{Hao.Zheng@ed.ac.uk}~~~~\texttt{mlap@inf.ed.ac.uk}\\
}

\makeatletter
\newcommand{\thickhline}{%
    \noalign {\ifnum 0=`}\fi \hrule height 1pt
    \futurelet \reserved@a \@xhline
}
\makeatother

\begin{document}
\maketitle
\begin{abstract}
  Compositional generalization is a basic mechanism in human language
  learning, which current neural networks struggle with.  A recently
  proposed \textbf{D}isent\textbf{angle}d sequence-to-sequence model
  (Dangle) shows promising generalization capability by learning
  specialized encodings for each decoding step. We introduce two key
  modifications to this model which encourage more disentangled
  representations and improve its compute and memory efficiency,
  allowing us to tackle compositional generalization in a more
  realistic setting.  Specifically, instead of adaptively re-encoding
  source keys and values at each time step, we disentangle their
  representations and only re-encode keys periodically, at some
  interval.  Our new architecture leads to better generalization
  performance across \emph{existing} tasks and datasets, and a \emph{new} machine
  translation benchmark which we create by
  detecting \emph{naturally occurring} compositional patterns in
  relation to a training set. We show this methodology  better emulates real-world
  requirements than artificial challenges.\footnote{Our code and
  dataset will be available at \url{xxx.yyy.zzz}.}

%Experimental results on the newly created
%  and two existing benchmarks (i.e.,semantic parsing and machine
%  translation) demonstrate that our proposal \mbox{R-Dangle} delivers better
%  compositional generalization and computational efficienty than
%  Dangle.  instances
\end{abstract}

\section{Introduction}

The Transformer architecture \cite{NIPS2017_3f5ee243} and variants
thereof have become ubiquitous in natural language processing. Despite
widespread adoption, there is mounting evidence that Transformers as
sequence transduction models struggle with \emph{compositional
  generalization} \cite{kim-linzen-2020-cogs, keysers2020measuring,
  li-etal-2021-compositional}. It is basically the ability to produce
and understand a potentially infinite number of novel linguistic
expressions by systematically combining known atomic components
\cite{Chomsky, montague1970universal}. Attempts to overcome this
limitation have explored various ways to explicitly inject
compositional bias through data augmentation
\cite{jia-liang-2016-data,Akyurek:ea:2020,andreas-2020-good,wang-etal-2021-learning-synthesize}
or new training objectives
\cite{conklin-etal-2021-meta,Oren:ea:2020,yin-etal-2021-compositional}. The majority of existing
approaches have been designed with semantic parsing in mind,  and as a result
adopt domain- and task-specific grammars or rules which do not
extend to other tasks (e.g.,~machine translation). 

% \cc{?} A similar
% philosophy underlies the creation of compositional generalization
% datasets, which \emph{synthesize} artificial test examples
% \cite{lake2018generalization,kim-linzen-2020-cogs,keysers2020measuring}
% or \emph{split} the training and test set in some heuristic or
% unrealistic way \cite{finegan-dollak-etal-2018-improving} which can be
% justified for semantic parsing but hardly applies to other
% sequence-to-sequence tasks.

%These approaches mostly exploit domain-specific grammars or rules designed for semantic parsing, hindering their applicability to other tasks (e.g., machine translation or NLG).  In stark contrast, developing general network architecture for compositional generalization has the potential to be applicable to a wide range sequence-to-sequence tasks.

In this work we aim to improve generalization via general
architectural modifications which are applicable to a wide range of
tasks. Our starting point are \citet{hao2022dangle} who unveil that
one of the reasons hindering compositional generalization in
Transformers relates to their representations being entangled. They
introduce {Dangle}, a sequence-to-sequence model, which learns more
\textbf{D}isent\textbf{angle}d representations by adaptively
re-encoding (at each time step) the source input. For each decoding
step, Dangle learns specialized source encodings by conditioning on
the newly decoded target which leads to better compositional
generalization compared to vanilla Transformers where source encodings
are shared throughout decoding. Although promising, their results are
based on synthetic datasets, leaving open the question of whether
{Dangle} is effective in real-world settings involving both complex
natural language and compositional generalization.

We present two key modifications to {Dangle} which
encourage learning \emph{more disentangled} representations \emph{more
efficiently}.  The need to perform re-encoding at each time step
substantially affects Dangle's training time and memory footprint. It
becomes prohibitively expensive on datasets with long target
sequences, e.g.,~programs with 400+ tokens in datasets like SMCalFlow
\cite{SMDataflow2020}. To alleviate this problem, instead of
adaptively re-encoding at each time step, we only re-encode
periodically, at some interval. Our decoder is no different from a
vanilla Transformer decoder except that it just re-encodes once in a
while in order to update its history information.  Our second
modification concerns disentangling the representations of source keys
and values, based on which the encoder in Dangle (and also in
Transformers) passes source information to the decoder. Instead of
computing keys and values using shared source encodings, we
disassociate their representations: we encode source \emph{values
once} and re-encode \emph{keys periodically}.

We evaluate the proposed model on existing benchmarks
\cite{SMDataflow2020,li-etal-2021-compositional} and a new  dataset which we create to better emulate a real-world
setting.  We develop a new methodology for \emph{detecting} examples
representative of compositional generalization in naturally occurring
text. Given a training and test set:  (a)~we discard examples from the test set that contain
out-of-vocabulary (OOV) or rare words (in relation to training) to
exclude novel atoms which are out of scope for compositional
generalization; (b)~we then measure how compositional a certain test
example is with respect to the training corpus; we introduce a metric
which allows us to identify a candidate pool of highly compositional
examples; (c)~using uncertainty estimation, we further select examples
from the pool that are both compositional in terms of surface form and
challenging in terms of generalization difficulty. Following these
three steps, we create a \emph{machine translation} benchmark using the IWSLT~2014 German-English
dataset as our training corpus and the WMT~2014 German-English
shared task as our test corpus. 

Experimental
results demonstrate that our new architecture achieves better
generalization performance across tasks and datasets and is adept at handling
real-world challenges.  Machine translation experiments on a diverse corpus of 1.3M WMT examples 
show it is particularly effective for long-tail compositional patterns. 

%%%%%%%%%%%%%%%%%%%%%%%%%%%%%%%%%%%%%%%%%%%%%%%%%%%%%%%%%%%%%
\section{Background: The {Dangle} Model}
\label{sec:dangle}

We first describe {Dangle}, the \textbf{D}isent\textbf{angle}d
Transformer model introduced in \citet{hao2022dangle} focusing on
their encoder-decoder architecture which they show delivers better
performance on complex tasks like machine translation.

Let $X=[x_1, x_2,...,x_n]$ denote a source sequence;
let~$f_\texttt{Encoder}$ and $f_\texttt{Decoder}$ denote a Transformer
encoder and decoder, respectively. $X$ is first encoded into a
sequence of contextualized representations~$N$:
\begin{eqnarray}
N & = & f_{\texttt{Encoder}}(X)
\end{eqnarray}
which are then used to decode target tokens $[y_1, y_2,...,y_m]$ one
by one. At the $t$-th decoding step, the Transformer takes~$y_t$ as input,
 reusing the source encodings~$N$ and  target memory~$M_{t-1}$
which contains the history hidden states of all decoder layers
corresponding to past tokens~$[y_1,y_2,...,y_{t-1}]$:
\begin{eqnarray}
y_{t+1}, M_t = f_\texttt{Decoder}(y_t, M_{t-1}, N)
\end{eqnarray}
This step not only generates a new token $y_{t+1}$, but also
updates the internal target memory $M_t$ by concatenating $M_{t-1}$
with the newly calculated hidden states corresponding to~$y_t$.

{Dangle} differs from vanilla Transformers in that it concatenates the
source input with the previously decoded target to construct
target-dependent input for \emph{adaptive} decoding:
\begin{eqnarray}
C_t & = & [x_1,x_2,...,x_n,y_1,...,y_t] \\
H_t & = & f_{\texttt{Adaptive\_Encoder}}(C_t)
\end{eqnarray}
The adaptive encoder consists of two components. $C_t$~is first fed to
$k_1$~Transformer encoder layers to fuse the target information:   
\begin{eqnarray}
\bar{H}_t = f_{\texttt{Adaptive\_Encoder}_1}(C_t)
\end{eqnarray}
where $\bar{H}_t$ is a sequence of contextualized representations
$[\bar{h}_{t,1},\bar{h}_{t,2},...,\bar{h}_{t,n},\bar{h}_{t,n+1},...,\bar{h}_{t,n+t}]$. Then,
the first~$n$ vectors corresponding to source tokens are extracted and
fed to another $k_2$~Transformer encoder layers for further
processing:
\begin{eqnarray}
% \bar{H}_{t,n} & = &[\bar{h}_{t,1},\bar{h}_{t,2},...,\bar{h}_{t,n}] \\
H_t & = & f_{\texttt{Adaptive\_Encoder}_2}(\bar{H}_t[:n])
\end{eqnarray}
Finally, the adaptive source encodings~$H_t$ together with the target
context~$[y_1,y_2,...,y_t]$ are fed to a Transformer decoder to
predict~$y_{t+1}$:
\begin{eqnarray}
y_{t+1}, M_t = f_\texttt{Decoder}(y_{<t+1}, \{\}, H_t)
\end{eqnarray}
In a departure from vanilla Transformers, {Dangle}
does not reuse the target memory from previous steps, but instead
re-computes \emph{all} target-side hidden states based on new
source encodings~$H_t$.

Similarly to Transformers, {Dangle} accesses source information
at each decoding step via encoder-decoder attention layers where the
same encodings~$H_t$ are used to compute both keys~$K_t$ and
values~$V_t$:
\begin{eqnarray}
K_t & = & H_tW^K \\ V_t & = & H_tW^V \\ O_t & = &
\mathrm{Attention}(Q_t, K_t, V_t)
\end{eqnarray}
where key and value projections~$W^K$ and~$W^V$ are parameter
matrices; and $Q_t$, $K_t$, $V_t$, and $O_t$ are respectively query,
key, value, and output matrices, at time step~$t$.
%%%%%%%%%%%%%%%%%%%%%%%%%%%%%%%%%%%%%%%%%%%%%%%%%%%%%%%%

\section{The \mbox{R-Dangle} Model}
\label{sec:modifications}

% The core design principle underlying {Dangle} is that learning
% specialized source encodings will encourage the model to zero in on
% concepts relevant to each prediction, thereby disentangling their
% contribution.  However, specialization comes at the cost of in-domain
% performance and computational efficiency. We empirically find that
% increasing Dangle's model size fails to match the in-domain gain
% achieved by an equally-sized Transformer model, and this could also
% negatively affect its compositional generalization.  Furthermore, the need to
% perform re-encoding (and also re-decoding) at each time step
% substantially increases training and inference cost, rendering the
% model computationally infeasible for some real-world language tasks
% with very long target sequences (e.g., in the region of hundreds of
% tokens). We thus adopt a more realistic approach, sharing
% representations where possible and only specializing them where
% necessary. We call our model \mbox{R-Dangle} as a shorthand for
% \textbf{R}eal-world \textbf{D}isent\textbf{angle}d Transformer.

In this section, we describe the proposed model, which we call  \mbox{R-Dangle} as a shorthand for \textbf{R}eal-world \textbf{D}isent\textbf{angle}d Transformer.

\subsection{Re-encoding at Intervals}
\label{sec:re-encoding-at}
The need to perform re-encoding (and also re-decoding) at each time
step substantially increases Dangle's training cost and memory
footprint, so that it becomes computationally infeasible for
real-world language tasks with very long target sequences (e.g., in
the region of hundreds of tokens).  Adaptively re-encoding at every
time step essentially means separating out relevant source concepts
for each prediction. However, the Transformer is largely capable of
encoding source phrases and decoding corresponding target phrases (or
logical form fragments in semantic parsing), as evidenced by its
remarkable success in many machine translation and semantic parsing
benchmarks
\cite{NIPS2017_3f5ee243,keysers2020measuring,zheng-lapata-2021-compositional-generalization}. This
entails that the entanglement problem (i.e., not being able to
disassociate the representations of different concepts for a sequence
of predictions) does not occur very frequently. We therefore relax the
strict constraint of re-encoding at every step in favor of the more
flexible strategy of re-encoding at intervals.

Given source sequence $X=[x_1, x_2,..., x_n]$, we
specify~$P=[t_1,t_2,...,t_l] (t_{i+1} - t_i =~o) $ in advance, i.e., a
sequence of re-encoding points with interval~$o$.  Then, during
decoding, when reaching a re-encoding point $t (t = t_i)$, we update
source encodings~$H_t$ and target memory~$M_t$:
\begin{eqnarray}
% H_{t_i} & = & f_{\texttt{Adaptive\_Encoder}}(C_{t_i}) \\
% y_{t_i+1}, M_{t_i}  & = & f_\texttt{Decoder}(y_{<t_i+1}, \{\}, H_{t_i}) \quad
H_t & = & f_{\texttt{Adaptive\_Encoder}}(C_t) \\
y_{t+1}, M_t  & = & f_\texttt{Decoder}(y_{<t+1}, \{\}, H_t) \quad
\end{eqnarray}
where $f_{\texttt{Adaptive\_Encoder}}$ denotes the adaptive encoder
described in Section~\ref{sec:dangle}.  For the next time step $t (
t_i < t < t_{i+1})$, we fall back to the vanilla Transformer decoder
using the source encodings~$H_{t_i}$ computed at time step~$t_i$:
\begin{eqnarray}
y_{t+1}, M_t  = f_\texttt{Decoder}(y_t, M_{t-1},  H_{t_i})
\end{eqnarray}
Note that we always set $t_1$ to~1 to perform adaptive encoding at the
first time step.

% \begin{eqnarray}
% y_{t+1}, M_t  & = & f_\texttt{Decoder}(y_{<t+1}, \{\}, H_{t,n}) \\
% & & \quad \quad \quad \quad\quad (\   \text{If } t \in P \ ) \nonumber \\
% y_{t+1}, M_t  & = & f_\texttt{Decoder}(y_t, M_{t-1}, N) \\
% & & \quad \quad \quad \quad\quad (\   \text{Otherwise} \ ) \nonumber \\
% y_{t+1}, M_t  = 
% \begin{cases}
%     f_\texttt{Decoder}(y_{<t+1}, \{\}, H_{t,n})    \\
%     \quad  \quad  \quad  \quad \quad \quad \quad \textbf{if } t \in P \\
%     f_\texttt{Decoder}(y_t, M_{t-1}, N) \\
%     \quad  \quad  \quad  \quad \quad \quad \quad \textbf{otherwise}
%  \end{cases} 
% \end{eqnarray}

%The WMT corpus provides a natural distribution shift compared to
%IWSLT, but most of problems in this setting may have no bearing on
%compositional generalization. We thus exploit the following three
%procedures to detect examples that maximally isolate compositional
%generalization.
\begin{table*}[t]
%\LARGE 
\centering
\begin{tabular}{cccc}
%{0.8cm}{5.4cm}{1.2cm}{0.8cm}}
\thickhline
\multicolumn{1}{c}{\bf Selected} &  \multicolumn{1}{c}{{\bf Examples}} & \multicolumn{1}{c}{{\bf Compositional Degree}} & \multicolumn{1}{c}{{\bf Uncertainty}} \\
\thickhline
\xmark  & \textcolor{cyan}{but} \textcolor{gray}{what can we do about this ?} & 2 / 8 = 0.25 & --- \\
\xmark  & \textcolor{gray}{please} \textcolor{pink}{report} \textcolor{darkgray}{all} \textcolor{cyan}{changes} \textcolor{violet}{here .} & 5 / 6 = 0.83 & 0.054 \\
\cmark & \textcolor{pink}{you have} \textcolor{violet}{disabled} \textcolor{gray}{your} \textcolor{orange}{javascript} ! & 5 / 6 = 0.83 & 0.274 \\ \hline

\thickhline 

\end{tabular}

\caption{Candidate examples from the WMT corpus. Different $n$-grams
  previously seen in the IWSTL training corpus are highlighted in
  color. The first 
  example is composed of two $n$-grams (\textsl{but} and \textsl{what
    can we do about this?}) with a compositional degree~0.25,
  and is discarded in the second stage. The second example has a high
  compositional degree but receives a low uncertainty score,
  and is thus
  filtered in the third stage. The third example is high in terms
  of both compositional degree and uncertainty, and is included in
  the compositional test set. } \label{table:system_output} 
\vspace{-0.2in}
\end{table*}

\subsection{Disentangling Keys and Values}
\label{sec:shar-valu-thro}
During decoding, {Dangle} accesses source information via
cross-attention (also known as encoder-decoder attention) layers where
the same source encodings are used to compute both keys \emph{and}
values.  The core design principle underlying Dangle is that learning
specialized representations for different purposes will encourage the
model to zero in on relevant concepts, thereby disentangling their
representations. Based on the same philosophy, we assume that source
keys and values encapsulate different aspects of source information,
and that learning more specialized representations for them would
further improve disentanglement, through the separation of the
concepts involved.
 
A straightforward way to implement this idea is using two separately
parameterized encoders to calculate two groups of source encodings
(i.e., corresponding to keys and values, respectively) during
re-encoding. However, in our preliminary experiments, we observed this
leads to serious overfitting and performance degradation. Instead, we
propose to encode values once and only update keys during adaptive
encoding. We compute source \emph{values} via the standard
Transformer encoder:
\begin{eqnarray}
H^v & = & f_{\texttt{Encoder}}(X)
\end{eqnarray}
and adaptively re-encode source \emph{keys} at an interval:
\begin{eqnarray}
H_t^k \mkern-13mu & = & \mkern-13mu f_{\texttt{Adaptive\_Encoder}}(C_t) \\
y_{t+1}, M_t \mkern-13mu & = & \mkern-13mu f_\texttt{KV\_Decoder}(y_{<t+1}, \{\}, H^v, H_t^k) \quad \quad
\end{eqnarray}
where $f_\texttt{KV\_Decoder}$ denotes a slightly modified Transformer
decoder where source keys and values in each cross-attention layer are
calculated based on different source encodings:
\begin{eqnarray}
K_t & = & H_t^kW^K \\
V & = & H^vW^V \\
O_t & = & \mathrm{Attention}(Q_t, K_t, V) 
\end{eqnarray}
At time step $t$ (where $t_i < t < t_{i+1}$), we perform vanilla
Transformer decoding:
\begin{eqnarray}
y_{t+1}, M_t \hspace*{-.1ex}=\hspace*{-.1ex}  f_\texttt{KV\_Decoder}(y_t, M_{t-1}, H^v, H_{t_i}^k) 
\end{eqnarray}

Note that fully sharing values could potentially cause some
entanglement, however, we we did not observe this in practice. We also
experimented with a variant where keys are shared and values are
repeatedly re-computed but empirically observed it obtains
significantly worse generalization performance than the value-sharing
architecture described above. This indicates that entanglement
 is more likely to occur when sharing keys.
 
%  , however, we leave
% further formal characterization of this behavior to future work.

%%%%%%%%%%%%%%%%%%%%%%%%%%%%%%%%%%%%%%%%%%%%%%%%%%%%%%%%%%%%%%%%%%%%%%%%%%
\section{A Real-world Compositional Generalization Challenge}
\label{sec:real-world-comp}

Models of compositional generalization are as good as the benchmarks
they are evaluated on. A few existing benchmarks are made of
artificially synthesized examples using a grammar or rules to systematically
control for different types of generalization
\cite{lake2018generalization,kim-linzen-2020-cogs,keysers2020measuring,li-etal-2021-compositional}. Unfortunately,
synthetic datasets lack the complexity of real natural language and
may lead to simplistic modeling solutions that do not generalize to
real world settings \cite{dankers-etal-2022-paradox}.  Other benchmarks  focus on
naturally occurring examples but have artificial train-test splits
\cite{finegan-dollak-etal-2018-improving}, based on heuristics
(e.g.,~query patterns).   Again, the types of compositional
generalization attested therein may not reflect real-world occurrence.

It is fair to assume that a SOTA model deployed in the wild, e.g.,~a
Transformer-based translation system, will be constantly presented
with new test examples. Many of them could be similar to seen training
instances or compositionally
different but in a way that does not pose serious generalization
challenges. An ideal benchmark for evaluating compositional
generalization should therefore consist of phenomena that are of
practical interest while challenging for SOTA models. To this end, we
create ReaCT, a new \textbf{REA}l-world dataset for
\textbf{C}ompositional generalization in machine
\textbf{T}ranslation. Our key idea is to obtain a generalization test
set by \emph{detecting} compositional patterns in relation to an
existing training set from a large and diverse pool of
candidates. Specifically, we use the IWSLT~2014 German $\to$ English
dataset as our training corpus and the WMT~2014 German $\to$ English
shared task as our test corpus (see Section~\ref{sec:experiments} for
details) and detect from the pool of WMT instances those that
exemplify compositional generalization with respect to IWSLT. This
procedure identifies naturally occurring compositional patterns which
we hope better represent practical generalization requirements than
artificially constructed challenges.

In the following, we describe how we identify examples that demand
compositional generalization.  While we create our new benchmark with
machine translation in mind, our methodology is general and applicable
to other settings such as semantic parsing. For instance, we could
take a relatively small set of annotated user queries as our training
set and create a generalization challenge from a large pool of
unlabeled user queries.

\paragraph{Filtering Out-of-Vocabulary Atoms} Compositional
generalization involves generalizing to \emph{new} compositions of
\emph{known} atoms. The WMT corpus includes many new semantic and
syntactic atoms that are not attested in IWSLT. A large number of
these are out-of-vocabulary (OOV) words which are by definition
unknown and out of scope for compositional generalization. We thus
discard WMT examples with words occurring less than 3~times in the
IWSLT training set which gives us approximately a pool of 1.3M
examples. For simplicity, we do not consider any other types of new
atoms such as unseen word senses or syntactic patterns.

\paragraph{Measuring Compositionality} How to define the notion of
compositional generalization is a central question in creating a
benchmark. Previous definitions have mostly centered around linguistic notions such as constituent or context-free grammars \cite{kim-linzen-2020-cogs,
  keysers2020measuring, li-etal-2021-compositional}. Since we do not
wish to synthesize artificial examples but rather detect them in
real-world utterances, relying on the notion of constituent might be
problematic. Sentences in the wild are often noisy and ungrammatical
and it is far from trivial to analyze their syntactic structure so as
to reliably identify new compositions of known constituents.  We overcome this problem by devising a metric based on
n-gram matching which assesses how compositional a certain example is
with respect to a training corpus.

Specifically, we first create a lookup dictionary of atomic units by
extracting all \mbox{$n$-grams} that occur more than 3~times in the
training corpus. Given a candidate sentence, we search the dictionary
for the minimum number of $n$-grams that can be composed to form the
sentence. For example, for sentence ``$x_1x_2x_3x_4x_5$'' and
dictionary $(x_1, x_2, x_3x_4, x_5, x_1x_2, x_3x_4x_5,)$, the minimum
set of such $n$-grams is $(x_1x_2, x_3x_4x_5)$. A sentence's
\emph{compositional degree} with respect to the training corpus is
defined as the ratio of the minimum number of $n$-grams to its length
(e.g.,~\mbox{2/5 = 0.4} for the above example). We select the
top~60,000 non-overlapping examples with the highest compositional
degree as our \emph{candidate pool}. As we discuss in
Section~\ref{sec:results}, compositional degree further allows us to
examine at a finer level of granularity how model performance changes
as test examples become increasingly compositional.

\paragraph{Estimating Uncertainty} Examples with the same
compositional degree could pose more or less difficulty to neural
sequence models (see last two utterances in Table~\ref{table:system_output}).
% (see the utterances \textsl{please report all changes
%   here.} and \textsl{you have disabled your javascript!} 
% in Table~\ref{table:system_output}). 
Ideally, we would like to identify
instances that are compositional in terms of surface form \emph{and}
hard in terms of the underlying generalization (see third example
in Table~\ref{table:system_output}). We detect such examples using a
metric  based on \emph{uncertainty estimation} and orthogonal
to compositional degree. We quantify predictive uncertainty based on
model ensembles, a method which has been successfully applied to
detecting misclassifications and out-of-distribution examples
\cite{NIPS2017_9ef2ed4b, malinin2021uncertainty}.

We follow the uncertainty estimation framework introduced in
\citet{malinin2021uncertainty} for sequence prediction tasks.
Specifically, we train 10~Transformer models with different random
initializations on IWSLT (our training corpus), and run inference over
the candidate pool created in the previous stage; for each example in
this pool, we measure the disagreement between ensemble models using
\emph{reverse mutual information}, a novel measure \cite{Malinin:2019,
  malinin2021uncertainty} which quantifies \emph{knowledge
uncertainty}, i.e., a model's uncertainty in its prediction due to
lack of understanding of the data rather than any intrinsic
uncertainty associated with the task (e.g.,~a word could have multiple
correct translations). We use the token-level approximation of
knowledge uncertainty.

We empirically find that the most uncertain examples are
extremely noisy and barely legible (e.g.,~they include abbreviations,
typos, and non-standard spelling). We thus  therefore throw away the top 2,000~uncertain examples and randomly sample 3,000~instances from
the next~18,000 most uncertain examples in an attempt to create a
generalization test set with diverse language patterns and different
levels of uncertainty.

%Overall, our
%compositional generalization test set contains 3,000 examples.
\begin{table}[t]
\centering
\setlength{\tabcolsep}{4pt} % Default value: 6pt

\scalebox{0.8}{
\begin{tabular}{@{}l|@{~}c@{~}c@{}r@{~~}r@{~~}rc@{}}
  \thickhline
  & & Comp & \multicolumn{2}{c}{Word $n$-gram} & \multicolumn{2}{c}{POS $n$-gram} \\
 Dataset                &\# examples  & Degree & \multicolumn{1}{c}{2} & \multicolumn{1}{c}{3} & \multicolumn{1}{c}{2} & \multicolumn{1}{c}{3} \\
                  \thickhline
                
                  COGS & 21,000 & 0.392 & 6,097 & 24,275 & 12 & 27 \\
                  CoGnition & 10,800 & 0.502 &  1,865  & 13,344 & 1 & 38 \\
                  CFQ & 11,968 & 0.268 & 168  & 2,736 & 8 & 30 \\
                 \thickhline
                  ReaCT & \hspace*{1.1ex}3,000 & 0.811 & 19,315 & 33,652  & 76 & \hspace*{-1.1ex}638 \\
                   \thickhline
  
\end{tabular}
}
\caption{Dataset Statistics: unique novel $n$-grams computed over
  words and parts of speech in ReaCT, and test partitions of COGS,
  CoGnition, and CFQ benchmarks. }
\label{tab:data}

\end{table}

\paragraph{Analysis}

We analyze the compositional nature of ReaCT by comparing it to
several popular benchmarks.  Specifically, for all datasets, we count
the number of novel test set $n$-grams that have not been seen in the
training.  We extract $n$-grams over words and parts of speech (POS);
word-based \mbox{$n$-grams} represent more superficial lexical composition
while $n$-grams based on POS tags reflect more of syntactic composition.

As shown in Table~\ref{tab:data}, despite being considerably smaller
compared to other benchmarks (see \#~examples column), ReaCT presents
substantially more diverse patterns in terms of lexical and syntactic
composition. It displays a much bigger number of novel word $n$-grams,
which is perhaps not surprising. Being a real-world dataset, it has a
larger vocabulary and more linguistic variation.  While our dataset
creation process does not explicitly target novel syntactic patterns
(approximated by POS $n$-grams), ReaCT still includes substantially
more compared to other benchmarks. This suggests that it captures the
complexity of real-world compositional generalization to a greater
extent than what is achieved when examples are synthesized
artificially. We show ReaCT examples with novel POS $n$-gram
compositions in  Appendix~\ref{sec:examples} (Table~\ref{table:example}).

%\textbf{Hao add an example here showing two sentences with same
%  compositional degree, but different uncertainty scores, also
%  illustrate token- versus sequence-level uncertainty}

\section{Experimental Setup}
\label{sec:experiments}
%In this section, we present our experimental setup for evaluating the
%proposed {R-Dangle} model.  We discuss the various datasets we used in
%our experiments, and implementation details regarding our model and
%comparison models.

\paragraph{Datasets}
We evaluated R-Dangle on two machine translation datasets and one
semantic parsing benchmark which we selected to maximally reflect
natural language variations and real-world generalization
challenges. These include: (a) \textbf{ReaCT}, the machine translation
benchmark developed in this paper; we used the IWSLT 2014 De$\to$En test set as the in-domain test set and create an out-of-distribution test set
from the WMT’14 De$\to$En training corpus; (b) \textbf{CoGnition}
\cite{li-etal-2021-compositional} is a semi-natural machine translation
benchmark focusing on English-Chinese sentence pairs; source sentences
were taken from the Story Cloze Test and ROCStories Corpora
\cite{mostafazadeh-etal-2016-corpus, mostafazadeh-etal-2017-lsdsem}
and target sentences were constructed by post-editing the output of a
machine translation engine; (c)~\textbf{SMCalFlow-CS}
\cite{SMDataflow2020} is a semantic parsing dataset for task-oriented
dialogue, featuring real-world human-generated utterances about
calendar management; following previous work
\cite{yin-etal-2021-compositional,qiu2021improving}, we report
experiments on the compositional skills split, considering a few-shot
learning scenario (with 6, 16, and~32 training examples). See
Appendix~\ref{sec:dataset_details} for more detail on datasets.

%We
%further describe below the parameterization of the models we used in
%our experiments.

\paragraph{Models}
On machine translation, our experiments evaluated two variants of \mbox{R-Dangle} depending on
whether keys and values are shared (\mbox{R-Dangle}$_{\rm shr}$) or
separate (\mbox{R-Dangle}$_{\rm sep}$).  We implemented all machine
translation models with fairseq \cite{ott2019fairseq}. We compared
\mbox{R-Dangle} against a vanilla Transformer
\cite{NIPS2017_3f5ee243} and the original Dangle model
\cite{hao2022dangle} which use the popular fairseq
configuration \texttt{transformer\_iwslt\_de\_en}.  We also implemented bigger variants of these
models using 12 encoder layers and 12 decoder layers which empirically
led to better performance. \mbox{R-Dangle}$_{\rm shr}$ and
\mbox{R-Dangle}$_{\rm sep}$ also use a 12-layer decoder. We tuned the
number of layers of the adaptive components ($k_1 = 2$ and $k_2 = 10$)
on the development set. For \mbox{R-Dangle}$_{\rm sep}$, we adopted a
10-layer value encoder and a 10-layer key encoder ($k_1 = 2$ and $k_2
= 8$), with the top 8~layers in the two encoders being shared. This
configuration produced 12~differently parametrized transformer encoder
layers, maintaining identical model size to comparison systems.

Previous work
\cite{qiu2021improving} has shown the advantage of
pre-trained models on the SMCalFlow-CS dataset.
For our semantic parsing experiments, we therefore built \mbox{R-Dangle} on top
of BART-large \cite{lewis-etal-2020-bart}.  We only report results with
\mbox{R-Dangle}$_{\rm shr}$ as the \mbox{R-Dangle}$_{\rm sep}$
architecture is not compatible with BART. We again set $k_1=2$ and
$k_2=10$. We provide more detail on model configurations in 
Appendix~\ref{sec:implementation_details}.

\begin{table}[t]
\centering
\scalebox{0.9}{
\begin{tabular}{l|cccc} \thickhline
CoGnition      & 1 & 2 & 4 & 8 \\ \hline 
\mbox{R-Dangle}$_{\rm shr}$ & 62.5 & 62.3 & 62.3 & 61.9 \\
\mbox{R-Dangle}$_{\rm sep}$ & 63.4 & 63.1 & 62.3 & 62.1 \\ \hline\hline
ReaCT  & 1 & 2 & 4 & 8 \\ \hline
\mbox{R-Dangle}$_{\rm shr}$ & 11.8 & 11.9 & 11.8 & 11.6 \\
\mbox{R-Dangle}$_{\rm sep}$ & 12.3 & 12.2 & 11.9 & 11.7 \\ \thickhline
\end{tabular}
}
\caption{BLEU score for \mbox{R-Dangle} variants (with different
  re-encoding intervals) on CoGnition and ReaCT compositional
  generalization test sets.  Note that \mbox{R-Dangle}$_{\rm shr}$
  with interval 1 is Dangle.}
\label{tab:mt_ablation}

\end{table}

\begin{table*}[t]
\centering
\setlength{\tabcolsep}{4pt} % Default value: 6pt

\scalebox{0.88}{
\begin{tabular}{l|cccc|ccc}
\thickhline
\multirow{2}{*}{Models}  & \multicolumn{4}{c|}{CoGnition} & \multicolumn{3}{c}{ReaCT} \\
                  &   $\downarrow~$ErrR$_{\mathrm{Inst}}$ &  $\downarrow$~ErrR$_{\mathrm{Aggr}}$ & $\uparrow$ ind-test & $\uparrow$cg-test & $\uparrow$IWSLT14 & $\uparrow$cg-test \\
                  \thickhline
                  Transformer \cite{hao2022dangle} & 30.5 & 63.8 &69.2 & 59.4  & 34.4 & 9.5 & \\
                 Dangle \cite{hao2022dangle} & 22.8 & 50.6 & 69.1 & 60.6  & --- & ---  \\
                  Transformer (our implementation) & 23.4 & 53.7 & 70.8 & 61.9     & 36.0 & 11.4  \\
                 Dangle (our implementation) & 19.7 & 47.0 & 70.6 & 62.5  & 36.1 & 11.8 \\
                 \thickhline
                 
                  \mbox{R-Dangle}$_{\rm sep}$ (interval = 1 ) & 16.0 & 42.1 & 70.7 & 63.4  & 36.0 & 12.3  \\ \thickhline
  
\end{tabular}
}
\caption{\textbf{Machine Translation Results:} we compare
  \mbox{R-Dangle} to baseline models on CoGnition and ReaCT. For
  CoGnition, we report instance-wise and aggregate compound
  translation error rates 
  (ErrR) on the compositional generalization test set (cg-test) and BLEU on both in-domain test set (ind-test) and cg-test. For ReaCT, we report BLEU on the in-domain IWSLT 2014 De$\rightarrow$En test set
  and the compositional generalization test set (cg-test) created in this
  paper. Results are averaged over 5 random runs on CoGnition and 3 random runs on ReaCT.} 
\label{tab:main_mt}

\end{table*}

%%%%%%%%%%%%%%%%%%%%%%%%%%%%%%%%%%%%%%%%%%%%%%%%%%%%%%%%%%%%%%%%%%
\section{Results}
\label{sec:results}

\paragraph{Disentangling Keys and Values Improves Generalization}

Table~\ref{tab:mt_ablation} reports the BLEU score
\cite{papineni-etal-2002-bleu} achieved by the two \mbox{R-Dangle}
variants on ReaCT and CoGnition, across different re-encoding
intervals.  \mbox{R-Dangle}$_{\rm sep}$ is consistently better than
\mbox{R-Dangle}$_{\rm shr}$ which confirms that representing keys and
values separately is beneficial. We also observe that smaller
intervals lead to better performance (we discuss this further later).

 Table~\ref{tab:main_mt} compares R-Dangle$_{\rm sep}$ (with
 interval~1) against baseline models.  In addition to BLUE, we report
 novel compound translation error rate, a  metric introduced in
 \citet{li-etal-2021-compositional} to quantify the extent to which
 novel compounds are mistranslated. We compute error rate over
 instances and an aggregate score over contexts.
 \mbox{R-Dangle}$_{\rm sep}$ delivers compositional generalization
 gains over Dangle and vanilla Transformer models (both in terms of
 BLEU \emph{and} compound translation error rate), even though their
 performance improves when adopting a larger \mbox{12-layer}
 network. \mbox{R-Dangle}$_{\rm sep}$ achieves a new state of the art
 on CoGnition (a gain of~0.9 BLEU points over Dangle and 1.5 BLEU
 points over the Transformer baseline).  \mbox{R-Dangle}$_{\rm sep}$
 fares similarly on ReaCT; it is significantly superior to the
 Transformer model by 0.9 BLEU points, and Dangle by 0.5 BLEU
 points. Moreover, improvements on compositional generalisation are not
 at the expense of in-domain performance (R-Dangle obtains
 similar performance to the Transformer and Dangle on the IWSLT2014
 in-domain test set).

 \begin{table}[t]
\centering

\scalebox{0.84}{
\begin{tabular}{l|ccc}
% \toprule
\thickhline

\bf{System} & 8-$\mathbb{C}$ & 16-$\mathbb{C}$ & 32-$\mathbb{C}$ \\
\thickhline

BERT2SEQ & ---  & 33.6 & 53.5 \\
BERT2SEQ+SS  & ---  & 46.8  & 61.7 \\
C2F  & --- & 40.6 & 54.6 \\
C2F$+$SS  & --- & 47.4  & 61.9 \\
T5 &  34.7 & 44.7 & 59.0 \\
T5$+$CSL & 51.6  & 61.4 & 70.4 \\
% \midrule
\thickhline
BART & 32.1 & 47.2  & 61.9 \\
$+$\mbox{R-Dangle}$_{\rm shr}$ (interval = 6) & 36.3 & 50.6 & 64.1 \\
% \bottomrule
\thickhline

\end{tabular}
}

\caption{{\bf Semantic Parsing Results:} we compare \mbox{R-Dangle} to
  various systems on SMCalFlow-CS. \mbox{*-$\mathbb{C}$}  denote different settings with 8, 16, and 32 cross-domain
  examples  added to the training set. Results for BERT and C2F models
  are  from \citet{yin-etal-2021-compositional}. Results for T5
  models are from \citet{qiu2021improving}. Results for BART and R-Dangle are averaged over 3 random runs.}
\label{tab:main_sp}

\end{table}

\begin{figure}[t]
  \centering
  \begin{tabular}{@{\hspace*{-1ex}}l@{}l}
    ~~~(a) & ~~~(b) \\
    \begin{tikzpicture}[scale=0.5]
\pgfplotsset{every axis legend/.style={anchor= west,draw=none,}
}

\begin{axis}[name=plot1, xlabel=Compositional Degree,colormap/blackwhite,
y tick label style={black},
legend style= {at={(0.35,0.925)}},
]
\addplot[smooth,mark=*,red]
coordinates{(0.853, 0.755) (0.772, 0.568) (0.74, 0.37) (0.717, 0.395) (0.7, 0.333) (0.687, 0.25) (0.669, 0.413) (0.667, 0.127) (0.657, 0.198) (0.645, 0.263) (0.637, 0.315) (0.629, 0.095) (0.624, 0.282) (0.617, 0.137) (0.611, 0.127) (0.601, 0.313) (0.6, -0.023) (0.592, 0.128) (0.586, 0.135) (0.581, 0.025) (0.574, 0.09) (0.571, 0.195) (0.566, 0.135) (0.56, 0.135) (0.556, 0.157) (0.551, 0.098) (0.545, 0.137) (0.54, 0.14) (0.536, 0.092) (0.531, -0.01) (0.526, 0.078) (0.52, 0.133) (0.508, 0.133) (0.5, 0.145) (0.5, 0.18) (0.5, -0.013) (0.5, 0.0) (0.487, 0.088) (0.476, 0.038) (0.468, 0.14) (0.459, 0.075) (0.45, 0.167) (0.44, 0.135) (0.427, 0.0) (0.409, 0.172) (0.381, 0.133) (0.322, 0.127)};
\addlegendentry{Difference in BLEU}
\end{axis}

\end{tikzpicture}
 & 
\begin{tikzpicture}[scale=0.505]
\pgfplotsset{every axis legend/.style={anchor= west,draw=none,}
}

\begin{axis}[name=plot1, xlabel=Interval,colormap/blackwhite,
y tick label style={blue},
legend style= {at={(0.35,0.925)}},
]
\addplot[smooth,mark=*,blue]
coordinates{(10,15.8) (20,5.69) (30,4.41) (40,3.76) (50,3.43)
   (70,2.97) (90,2.73) (110,2.61) (130,2.54) (150,2.39) (170,2.34)};
\addlegendentry{Training time (hours)}
\end{axis}

\begin{axis}[name=plot2, axis y line*=right, axis x line=none,
 xlabel=Interval,colormap/blackwhite, 
y tick label style={red},
legend style= {at={(0.35,0.85)}},
]
\addplot[smooth,mark=square*,red]
coordinates{(10,50) (20,48.9) (30,48.1) (40,50.3) (50,49.7) (70,48) (90,46.3)
  (110,48.8) (130,47.6) (150,48.1) (170,47.8)};
\addlegendentry{Accuracy (\%)}
\end{axis}
\end{tikzpicture} 
\end{tabular}
\vspace{-1ex}
\caption{(a)~Difference in BLEU score between R-Dangle$_{\rm sep}$
    (interval = 1) and Transformer vs compositional degree. A positive
    score means R-Dangle$_{\rm sep}$ is better than Transformer. Each
    data point is computed on 30K WMT examples. R-Dangle shows
    increasing performance improvements as test examples become more
    compositional. (b)~Training cost (hours) and test accuracy vs interval
  length. \mbox{R-Dangle$_{\rm shr}$} was trained on SMCalFlow-CS (16-$\mathbb{C}$) using 4 A100 GPUs.}
\label{fig:comp_degree}
\vspace{-2ex}
\end{figure}
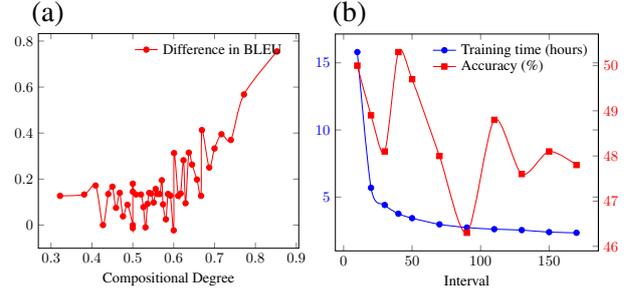

 \paragraph{R-Dangle Can Handle Long-tail Compositional Patterns Bettter}
We next examine model performance on  real-world examples with diverse language  and different levels of composition. 
Specifically, we
train \mbox{R-Dangle}$_{\rm sep}$
 (interval=1) and a Transformer  on the IWSTL14 corpus and test on the pool of 1.3M WMT examples 
 obtained after filtering OOV words.
Figure~\ref{fig:comp_degree}a plots
 the difference in BLEU between the two models against compositional degree. This
 fine-grained evaluation reveals that they perform similarly on
 the majority of less compositional examples (BLUE difference is
 around zero), however, the performance gap becomes larger with more
 compositional examples (higher difference means higher BLEU for
 \mbox{R-Dangle}$_{\rm sep}$). This indicates that R-Dangle is particularly effective for handling long-tail compositional patterns. 
 
 % We further examined where improvements in compositional
 % generalization are coming from.
 
 % Figure~\ref{fig:comp_degree}a plots
 % the difference in BLEU between \mbox{R-Dangle}$_{\rm sep}$
 % (interval=1) and Transformer against compositional degree. Both
 % models are trained on the IWSTL14 De$\to$En corpus and tested on the
 % pool of 1.3M WMT examples selected after filtering OOV words.  This
 % fine-grained evaluation reveals that both models perform similarly on
 % the majority of less compositional examples (BLUE difference is
 % around zero), however the performance gap is larger with more
 % compositional examples (higher difference means higher BLEU for
 % \mbox{R-Dangle}$_{\rm sep}$).

% different compositional region. Table \ref{fig:comp_degree} presents
% the results. Interestingly, the
% compositional test examples, the performance improvement from
% R-Dangle becomes more and more noticeable.

\paragraph{\mbox{R-Dangle} Boosts the Performance of Pretrained Models}
The ``pre-train and fine-tune'' paradigm \cite{peters-etal-2018-deep,
  devlin-etal-2019-bert, Raffel2020ExploringTL, lewis-etal-2020-bart}
has been widely adopted in NLP, and semantic parsing is no exception
\cite{shin-etal-2021-constrained, qiu2021improving}. We further
investigate \mbox{R-Dangle}'s performance when combined with a
pre-trained model on the SMCalFlow-CS dataset (across the three
cross-domain settings). Table~\ref{tab:main_sp} shows that
R-Dangle$_{\rm shr}$ boosts the performance of BART-large, which
suggests that generalization improvements brought by \mbox{R-Dangle}
are complementary to generalization benefits afforded by large-scale
pre-training (see \citealt{hao2022dangle} for a similar
conclusion). The proposed model effectively marries pre-training with
disentangled representation learning to achieve better generalization.

%\paragraph{How does \mbox{R-Dangle} Perform on Real-world Semantic
%  Parsing Data?}

In Table~\ref{tab:main_sp}, we also compare \mbox{R-Dangle} with other
top-performing models on SMCalFlow-CS. These include: (a)~a
sequence-to-sequence model with a BERT encoder and an LSTM decoder
using a copy mechanism (BERT2SEQ;
\citealt{yin-etal-2021-compositional}); (b)~the coarse-to-fine model
of \citet{dong-lapata-2018-coarse} which uses a BERT encoder and a
structured decoder that factorizes the generation of a program into
sketch and value predictions; (c) and combinations of these two models
with span-supervised attention ($+$SS;
\citealt{yin-etal-2021-compositional}). We also include a T5 model and
variant thereof trained on additional data using a model called
Compositional Structure Learner (CSL) to generate examples for data
augmentation (T5+CSL; \citealt{qiu2021improving}). \mbox{R-Dangle}
with BART performs best among models that do \emph{not} use data
augmentation across compositional settings. Note that our proposal is
orthogonal to CSL and could also benefit from data augmentation.

%\begin{figure}[t]
%  \centering
%\begin{tikzpicture}[scale=0.6]
%\pgfplotsset{every axis legend/.style={anchor= west,draw=none,}
%}
%
%\begin{axis}[name=plot1, xlabel=Interval,colormap/blackwhite,
%y tick label style={blue},
%legend style= {at={(0.35,0.925)}},
%]
%\addplot[smooth,mark=*,blue]
%coordinates{(10,15.8) (20,5.69) (30,4.41) (40,3.76) (50,3.43)
%   (70,2.97) (90,2.73) (110,2.61) (130,2.54) (150,2.39) (170,2.34)};
%\addlegendentry{Training time (hours)}
%\end{axis}
%
%\begin{axis}[name=plot2, axis y line*=right, axis x line=none,
% xlabel=Interval,colormap/blackwhite, 
%y tick label style={red},
%legend style= {at={(0.35,0.85)}},
%]
%\addplot[smooth,mark=square*,red]
%coordinates{(10,50) (20,48.9) (30,48.1) (40,50.3) (50,49.7) (70,48) (90,46.3)
%  (110,48.8) (130,47.6) (150,48.1) (170,47.8)};
%\addlegendentry{Accuracy (\%)}
%\end{axis}
%\end{tikzpicture}
%\vspace{-1ex}
%\caption{Training cost (hours) and test accuracy vs interval
%  length. \mbox{R-Dangle$_{k}$} was trained on SMCalFlow-CS using 4 A100 GPUs.}
%\label{fig:interval}
%\vspace{-2ex}
%\end{figure}

%In contrast, when decoding formal languages in semantic parsing could boost
%generalization performance with a much larger re-encoding interval
%(and thus larger speedup and smaller memory footprint).

\paragraph{Larger Re-encoding Intervals Reduce Training Cost}
%We further examine how different re-encoding intervals influence
%training time and generalization performance.  
The results in
Table~\ref{tab:mt_ablation} indicate that re-encoding correlates with
R-Dangle's generalization ability, at least for machine
translation. Both model variants experience a drop in BLEU points when
increasing the re-encoding interval to~8.  We hypothesize that this
sensitivity to interval length is task-related; target sequences in
machine translation are relatively short and representative of real
language, whereas in SMCalFlow-CS, the average length of target sequences (in formal language) is~99.5 and
the maximum length is~411.  It is computationally infeasible to train
\mbox{R-Dangle} with small intervals on this dataset, however, larger
intervals still produce significant performance gains.

Figure~\ref{fig:comp_degree}b shows how accuracy and training time vary
with interval length on SMCalFlow-CS with the 16-$\mathbb{C}$ setting. Larger
intervals substantially reduce training cost with an optimal
speed-accuracy trade off in between 10 and~50. For instance, interval
40 yields a 4x speed-up compared to interval 10 while achieving
50.3\%~accuracy. Finding a trade-off between generalization and
efficiency is an open research problem which we leave to future work.

%The optimal interval for \mbox{R-Dangle}$_{kv}$ is 2
%and 2 on CoGnition and ReaCT respectively; the optimal interval for
%\mbox{R-Dangle}$_{k}$ is 1 and 2 on CoGnition and ReaCT respectively.

%We may adopt some more
%flexible re-encoding strategy to obtain better efficiency and
%performance. We leave that to future work.

% \begin{itemize}
% \item Add in table 2 in-domain performance for CoGnition
% \item Graph with different interval $o$ values
% \item Mention which optimal $o$ for table 2 and Table 3 for R-dangle.
% \item Inspect examples that improve with R-Dangle vs Dangle
% \item Add in Appendix dataset statistics and examples for different
%   datasets. 
% \end{itemize}

\section{Related Work}

The realization that neural sequence-to-sequence models struggle with
compositional generalization has led to numerous research efforts
aiming to precisely define this problem and explore possible solutions
to it. A line of research focuses on benchmarks which capture
different aspects of compositional generalization.
\citet{finegan-dollak-etal-2018-improving} repurpose existing semantic
parsing benchmarks for compositional generalization by creating more
challenging splits based on logical form patterns.  In \textsc{SCAN}
\cite{DBLP:conf/icml/LakeB18} compositional generalization is
represented by unseen combinations of seen actions (e.g., JUMP
LTURN). \citet{keysers2020measuring} define compositional
generalization as generalizing to examples with maximum compound
divergence (e.g., combinations of entities and relations) while
guaranteeing similar atom distribution to the training set.
\citet{kim-linzen-2020-cogs} design five linguistic types of
compositional generalization such as generalizing phrase nesting to
unseen depths. In ReaCT, our definition of compositional
generalization is dependent on the data distribution of the candidate
corpus, which determines what compositional patterns are of practical
interest and how frequently they occur.

Another line of work focuses on modeling solutions, mostly ways to
explicitly instil compositional bias into neural models.  This can be
achieved by adopting a more conventional grammar-based approach
\cite{herzig2020spanbased} or incorporating a lexicon or lexicon-style
alignments into sequence models
\cite{akyurek-andreas-2021-lexicon,zheng-lapata-2021-compositional-generalization}. Other
work employs heuristics, grammars, and generative models to synthesize
examples for data augmentation
\cite{jia-liang-2016-data,Akyurek:ea:2020,andreas-2020-good,wang-etal-2021-learning-synthesize,
  qiu2021improving} or augments standard training objectives with new
supervision signals like attention supervision or meta-learning
\cite{Oren:ea:2020,conklin-etal-2021-meta,yin-etal-2021-compositional}. Our
work builds on {Dangle} \cite{hao2022dangle}, a disentangled sequence-to-sequence model,
which tries to tackle compositional generalization with architectural
innovations. While Dangle is conceptually general, our proposal is tailored to the Transformer and features two key modifications to
encourage more disentangled representations and better computational efficiency.

\section{Conclusions}

In this paper we focused on two issues related to compositional
generalization. Firstly, we improve upon Dangle, an existing
sequence-to-sequence architecture which generalizes to unseen
compositions by learning specialized encodings for each decoding
step. We show that re-encoding keys periodically, at some interval,
improves both efficiency and accuracy. Secondly, we propose a
methodology for identifying compositional patterns in real-world data
and create a new dataset which better represents practical
generalization requirements. Experimental results show that our
modifications improve generalization across tasks, metrics, and
datasets and our new benchmark provides a challenging testbed for
evaluating new modeling efforts.

\section*{Limitations}
On machine translation, the optimal generalization performance requires using small interval values. However, R-dangle with small intervals still runs much slower than an equivalent Transformer model.  In this paper, we only explore a simple periodic re-encoding strategy. However, more complex and flexible ways of re-encoding could be used to further narrow the gap. For instance, we could adopt a dynamic strategy which \emph{learns} when re-encoding is necessary.

\bibliography{anthology,custom}
\bibliographystyle{acl_natbib}

\appendix

\begin{table*}[t]
%\LARGE

\begin{small}

\begin{tabular}{m{0.59\linewidth}m{0.35\linewidth}}

\thickhline

 \multicolumn{1}{c}{Train} & \multicolumn{1}{c}{Test } \\  \thickhline
 \textbullet \hspace{1ex} and i can 't believe you 're here and that i 'm meeting \textcolor{orange}{you here at ted .} \hspace*{.3cm}( \cmtt{PRP RB IN NN .} ) \vspace{0.2cm} \hfill\break 
\textbullet \hspace{1ex} you see , this is what india is today . \textcolor{cyan}{the ground reality is based on} \hspace*{.489cm}(~\cmtt{DT NN NN VBZ VBN IN }) a cyclical world view . 
 &  \textcolor{cyan}{the account data is provided to} \textcolor{orange}{you directly via e-mail .}
 \\ \hline
 
 \textbullet \hspace{1ex} \textcolor{orange}{a couple of hours} ( \cmtt{DT NN IN NNS} ) later , the sun will shine on the next magnifying glass . \vspace{0.2cm} \hfill\break
 \textbullet \hspace{1ex} but this \textcolor{cyan}{could also be used for good .} ( \cmtt{MD RB VB VBN IN NN .} ) &
 \textcolor{orange}{both setting of tasks} \textcolor{cyan}{must successfully be mastered under supervision .}
 \\ \hline

 \textbullet \hspace{1ex} the \textcolor{orange}{national science foundation , other countries} ( \cmtt{JJ NN NN , JJ NN} ) are very interested in doing this
 \vspace{0.2cm} \hfill\break
 \textbullet \hspace{1ex} no , they \textcolor{cyan}{are full of misery .} ( \cmtt{VBP JJ IN NN .})  &
 its \textcolor{orange}{warm water temperature , small depth} \textcolor{cyan}{are convenient for bathing .}
 \\ \thickhline

\end{tabular}

 \end{small}

\caption{Novel syntactic compositions in ReaCT test set (syntactic
  atoms of same type are color coded). \mbox{POS-tag} sequences for these
  atoms are shown in parentheses (\cmtt{PRP:}pronoun,
  \cmtt{RB:}adverb, \cmtt{IN:} preposition, \cmtt{NN/S:}noun singular/plural, \cmtt{DT:}
  determiner, \cmtt{JJ:}
  adjective, \cmtt{MD:}modal,
  \cmtt{VBZ/P:} non-/3rd person singular present, \cmtt{VBN:} verb
  past participle.)} \label{table:example}

\vspace{-0.2in}
\end{table*}

\section{Examples from ReaCT Dataset}
\label{sec:examples}

Table~\ref{table:example} showcases examples from the ReaCT test
set. These are novel syntactic patterns approximated by POS
$n$-grams. As mentioned in Section~\ref{sec:real-world-comp}, ReaCT is
created by detecting compositional patterns in relation to an existing
training set from a diverse pool of candidates.

\section{Dataset Details}
\label{sec:dataset_details}

We evaluated our model on two machine translation datasets, and one
semantic parsing benchmark which we selected to maximally reflect
natural language variations and real-world generalization
challenges. We describe these in detail below.

\paragraph{ReaCT} \hspace*{-.25cm}is the real-world machine
translation benchmark developed in this paper for compositional
generalization.  The IWSLT 2014 De$\to$En dataset consists of
approximately 170K~sequence pairs. We used the fairseq script
\texttt{prepare-iwslt14.sh} to randomly sample approximately~4\% of
this dataset as validation set and kept the rest as training set.
Following standard practice, we created an in-domain test set, the
concatenation of files dev2010, dev2012, tst2010, tst2011, and
tst2012. We created an out-of-distribution test sets from the WMT’14
De$\to$En training corpus following the uncertainty selection method
based on sequences.

\paragraph{CoGnition}  \hspace*{-.25cm}is another machine translation
benchmark targeting compositional generalization
\cite{li-etal-2021-compositional}.  It also contains a synthetic test
set to quantify and analyze compositional generalization of neural MT
models.  This test set was constructed by embedding synthesized novel
compounds into training sentence templates.  Each compound was
combined with 5 different sentence templates, so that every compound
can be evaluated under 5 different contexts.  A major difference
between \textsc{ReaCT} and CoGnition is the fact that test sentences
for the latter are not naturally occurring. Despite being somewhat
artificial, CoGnition overall constitutes a realistic benchmark which
can help distinguish subtle model differences compared to purely
synthetic benchmarks. For example, \citet{hao2022dangle} showed that
their encoder-only Dangle variant performed badly on this dataset in
spite of impressive performance on synthetic semantic parsing
benchmarks \cite{kim-linzen-2020-cogs,keysers2020measuring}.

\paragraph{SMCalFlow-CS} \hspace*{-.25cm}\cite{SMDataflow2020} is a
large-scale semantic parsing dataset for task-oriented dialogue,
featuring real-world human-generated utterances about calendar
management.  \citet{yin-etal-2021-compositional} proposed a
compositional skills split of SMCalFlow (SMCalFlow-CS) that contains
single-turn sentences from one of two domains related to creating
calendar events (e.g.,~\textsl{Set up a meeting with Adam}) or
querying an org chart (e.g., \textsl{Who are in Adam's team?}), paired
with LISP programs. The training set $\mathbb{S}$ consists of samples
from single domains while the test set $\mathbb{C}$ contains
compositions thereof (e.g.,~\textsl{create a meeting with \uline{Adam
    and his team}}).  Since zero-shot compositional generalization is
highly non-trivial due to novel language patterns and program
structures, we follow previous work
\cite{yin-etal-2021-compositional,qiu2021improving} and consider a
few-shot learning scenario, where a small number of cross-domain
examples are included in the training set. We report experiments with
6, 16, and 32 examples.

\section{Implementation Details}
\label{sec:implementation_details}

\paragraph{Machine Translation Models} We implemented all translation models with fairseq \cite{ott2019fairseq}.  
Following previous work
\cite{li-etal-2021-compositional,hao2022dangle}, we compared with the
baseline machine translation models Dangle and Transformer using the
popular fairseq configuration \texttt{transformer\_iwslt\_de\_en}. We
also implemented a bigger variant of these models using a new
configuration, which empirically obtained better performance. We used
12 encoder layers and 12 decoder layers. We set the dropout to~0.3 for
attention weights and 0.4~after activations in the feed-forward
network. We also used pre-normalization (i.e.,~we added layer
normalization before each block) to ease optimization. Following
\citet{hao2022dangle}, we used relative position embeddings
\cite{shaw-etal-2018-self,huang-etal-2020-improve} which have
demonstrated better generalization performance.

Hyperparameters for \mbox{R-Dangle} were tuned on the respective
validation sets of CoGnition and ReaCT.  Both \mbox{R-Dangle}$_{\rm
  shr}$ and \mbox{R-Dangle}$_{\rm sep}$ used a 12-layer decoder. For
\mbox{R-Dangle}$_{\rm shr}$, we tuned the number of layers of the two
adaptive components $k_1$ and $k_2$, and set $k_1$ and $k_2$ to 2 and
10, respectively. For \mbox{R-Dangle}$_{\rm sep}$, we shared some
layers of parameters between the value encoder and the adaptive key
decoder and experimented with different sharing strategies. Finally,
we adopted a 10-layer value encoder and a 10-layer key encoder ($k_1 =
2$ and $k_2 = 8$). The top 8 layers in the two encoders were
shared. This configuration produced 12 differently parametrized
transformer encoder layers, thus maintaining identical model size to
the baseline.

\paragraph{Semantic Parsing Models}
\citet{qiu2021improving} showed the advantage of pre-trained
sequence-to-sequence models on SMCalFlow-CS. We therefore built
\mbox{R-Dangle} on top of BART-large \cite{lewis-etal-2020-bart},
which is well supported by fairseq. We used BART's encoder and decoder
to instantiate the adaptive encoder and decoder in our model. For
compatibility, we only employ the \mbox{R-Dangle}$_{\rm shr}$
architecture. We also set $k_1$ and $k_2$ to 2 and 10, respectively.

\end{document}